\documentclass[12pt,letterpaper]{article}

\usepackage{enumerate}
\usepackage{xparse}
\usepackage{array}
\usepackage{hhline}
\usepackage{makecell}
\usepackage{mathrsfs}
\usepackage[authoryear]{natbib}
\usepackage{extarrows}
\usepackage{graphicx,setspace,lscape,longtable}
\usepackage{mathrsfs,amsmath,amsthm,amssymb,color}
\usepackage{tabularx}
\usepackage{epsfig,graphicx,pdfpages}
\usepackage{rotating}
\usepackage{float}
\usepackage{bm}
\usepackage{ragged2e}
\usepackage{bbm}
\usepackage{hyperref}
\usepackage{multirow}
\usepackage[hmargin=0.9in,vmargin=1in]{geometry}
\newcommand{\blind}{0}
\newcommand{\spacingset}[1]{\setstretch{#1}\normalsize}
\hypersetup{
	colorlinks=true,
	linkcolor=blue,
	citecolor=blue,
	urlcolor=blue}
\usepackage{booktabs}
\usepackage[T1]{fontenc}
\usepackage[utf8]{inputenc}
\usepackage{enumitem}
\usepackage{algorithm}
\usepackage{algorithmic}
\usepackage{amsmath}
\usepackage{graphicx}
\usepackage{subfigure}
\def\be{\begin{eqnarray}}
	\def\ee{\end{eqnarray}}
\bibpunct{(}{)}{;}{a}{}{,}

\makeatletter
\DeclareRobustCommand{\thmprime}{%
  \begingroup
  \expandafter\in@\expandafter b\expandafter{\f@series}%
  \ifin@ \boldmath \fi
  $\m@th{}^{\prime}$%
  \endgroup
}
\makeatother

\newtheorem{theorem}{Theorem}
\newtheorem{lemma}{Lemma}
\newtheorem{proposition}{Proposition}

\newtheorem{assumption}{Condition}
\newtheorem{remark}{Remark}
\newtheorem{corollary}{Corollary}

\def\beq{\begin{equation}}
	\def\eeq{\end{equation}}
\def\beqr{\begin{eqnarray}}
	\def\eeqr{\end{eqnarray}}
\def\beqrs{\begin{eqnarray*}}
	\def\eeqrs{\end{eqnarray*}}
\def\bet{\begin{theorem}}
	\def\eet{\end{theorem}}
\def\bel{\begin{lemma}}
	\def\eel{\end{lemma}}
\def\bep{\begin{proposition}}
	\def\eep{\end{proposition}}
\def\bg{\begin{figure}[tbph]\begin{center}}
		\def\eg{\end{center}\end{figure}}

\def\bc{\begin{center}}
	\def\ec{\end{center}}

\NewDocumentEnvironment{manual}{O{theorem}m}
 {%
  \addtocounter{theorem}{-1}%
  \renewcommand\thetheorem{#2}%
  \begin{#1}%
 }
 {\end{#1}}%

\NewDocumentEnvironment{manuallemma}{m}
 {%
  \addtocounter{lemma}{-1}%
  \renewcommand\thelemma{#1}%
  \begin{lemma}%
 }
 {\end{lemma}}%

\NewDocumentEnvironment{manualprop}{m}
 {%
  \addtocounter{proposition}{-1}%
  \renewcommand\theproposition{#1}%
  \begin{proposition}%
 }
 {\end{proposition}}%

\NewDocumentEnvironment{manualassumption}{m}
 {%
  \addtocounter{assumption}{-1}%
  \renewcommand\theassumption{#1}%
  \begin{assumption}%
 }
 {\end{assumption}}%

\NewDocumentEnvironment{manualcorollary}{m}
 {%
  \addtocounter{corollary}{-1}%
  \renewcommand\thecorollary{#1}%
  \begin{corollary}%
 }
 {\end{corollary}}%

\def\to{\rightarrow}

\def\wt{\widetilde}
\def\wh{\widehat}

\def\mA{\mathcal{A}}

\def\mg{\mathbb{g}}

\def\mL{\mathcal{L}}

\def\mM{\mathbb{M}}

\def\mP{\mathcal{P}}

\def\mR{\mathbb{R}}

\def\mS{\mathcal{S}}

\def\argmin{\mbox{argmin}}

\newcommand{\RNum}[1]{\uppercase\expandafter{\romannumeral #1\relax}}

\numberwithin{equation}{section}
\allowdisplaybreaks[4] %

\def\SEW{\operatorname{SE}(W)}
\def\wt{\widetilde}

\newcommand{\Mean}{{\mathbb{E}}}

\def\omegaavg{\overline{\gamma}^{\scriptscriptstyle{\mA}}_2}
\def\flatavg{\overline{\flat}^{\scriptscriptstyle{\mA}}_2}
\def\beq{\begin{equation}}
\def\eeq{\end{equation}}
\def\beqr{\begin{eqnarray}}
\def\eeqr{\end{eqnarray}}
\def\beqrs{\begin{eqnarray*}}
\def\eeqrs{\end{eqnarray*}}

\def\mg{\mathcal G}

\usepackage{placeins,etoolbox}
\providecommand{\blue}{}

\begin{document}
\title{\bfseries Adaptive Decentralized Federated Learning for Robust Optimization}
\date{}
\ifnum\blind=0
\author{\begin{minipage}{\dimexpr\textwidth-2\tabcolsep\relax}
\centering\normalsize
Shuyuan Wu$^{1,*}$, Feifei Wang$^{2,*}$, Yuan Gao$^{3,\dagger}$, and Hansheng Wang$^{4}$\\[1ex]
{\small\itshape
$^1$School of Statistics and Data Science,\\
Shanghai University of Finance and Economics, China\\[0.4ex]
$^2$School of Statistics, Renmin University of China, China\\[0.4ex]
$^3$School of Statistics and Data Science,\\
Shanghai University of International Business and Economics, China\\[0.4ex]
$^4$Guanghua School of Management, Peking University, China}
\end{minipage}}
\else
\author{}
\fi

\spacingset{1}
\maketitle
\ifnum\blind=0
\footnotetext[1]{Shuyuan Wu and Feifei Wang are joint first authors.}
\footnotetext[2]{Yuan Gao is the corresponding author. Correspondence to: ygao\_stat@outlook.com.}
\fi
\begin{abstract}
\normalsize
In decentralized federated learning (DFL), the presence of abnormal clients, often caused by noisy or poisoned data, can significantly disrupt the learning process and degrade the overall robustness of the model. 
Previous methods on this issue often require a sufficiently large number of normal neighboring clients or prior knowledge of reliable clients, which reduces the practical applicability of DFL. 
To address these limitations, we develop here a novel adaptive DFL (aDFL) approach for robust estimation. The key idea is to adaptively adjust the learning rates of clients.
By assigning smaller rates to suspicious clients and larger rates to normal clients, aDFL mitigates the negative impact of abnormal clients on the global model in a fully adaptive way. Our theory does not put any stringent conditions on neighboring nodes and requires no prior knowledge.  
A rigorous convergence analysis is provided to guarantee the oracle property of aDFL. {We also propose a pessimistic decentralized cross-validation method to select the tuning parameter.} Extensive numerical experiments demonstrate the superior performance of the aDFL method.
\end{abstract}

\section{Introduction}
\label{sec:intro}
{\blue Distributed learning is an effective way to handle large-scale datasets by distributing computation across multiple clients \citep{he2024ddac}. When data are stored locally across multiple sites, decentralized federated learning (DFL) provides a practical framework for model training without pooling raw data \citep{beltran2023decentralized}; similar data-sharing constraints also arise in multicenter studies, where sequential summary-based inference has been developed \citep{hu2024collaborative}.}
Different from centralized federated learning (CFL), DFL eliminates the need for a central server and enables clients to collaborate 
in a peer-to-peer manner. {\blue This architecture reduces communication bottlenecks and single-point-of-failure risks.} Nevertheless, DFL also faces a serious challenge. 
DFL systems may involve unreliable or undesirable clients that degrade the overall performance of the learning process. One typical example is Byzantine failure, which refers to transmitting incorrect information to the whole network \citep{blanchard2017machine,yin2018byzantine,tu2021variance}. In addition, some clients may hold low-quality or even corrupted data, contain information irrelevant to the target task, exhibit severe distribution shifts, or suffer from unstable communication and computation. These issues could negatively impact the training process by introducing noise or bias into the model updates, which is especially problematic for DFL, as the lack of a central server makes it more difficult to detect and correct abnormal behaviors \citep{wu2023byzantine,zhang2024119808high}.

To address the abnormal clients in DFL, various robust learning methods have been proposed{\blue  . These methods mainly fall into two categories. Methods in the first category replace the simple average of neighboring gradients or parameter estimates with a robust aggregator, such as the coordinate-wise median, trimmed mean, or clipped average \citep{fang2022bridge,he2022selfcentered,wu2023byzantine}. Methods in the second category first identify abnormal clients through statistical testing and then remove them from the network \citep{zhang2026exact}. See Section~\ref{sec:review} for a more detailed discussion.}
{\blue However, these methods still suffer from several practical limitations. First, many of them require each client to have a sufficiently large number of normal neighbors; otherwise, robust summaries (e.g., the median) of neighboring gradients or parameter estimates cannot be reliably computed \citep{yang2019byrdie,su2020byzantine,fang2022bridge}. Second, some methods require prior knowledge of an upper bound on the number of abnormal neighbors, which is often unavailable in practice \citep{wu2023byzantine}. In addition, the identification-based method of \citet{zhang2026exact} requires the subgraph formed by normal clients to be connected.}

\begin{figure}[!t]
\centering
\includegraphics[width=0.92\textwidth]{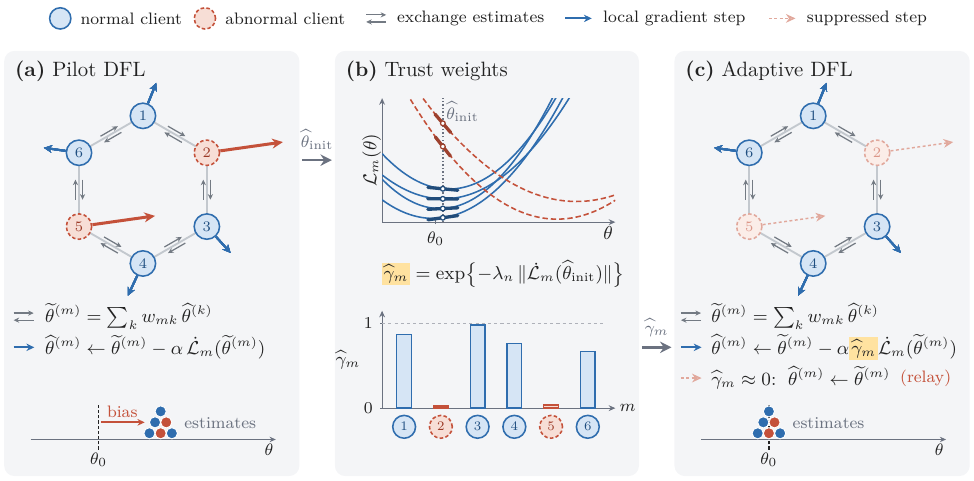}
\caption{Overview of aDFL with $M=6$ clients; clients 2 and 5 (dashed) are abnormal. Clients only exchange estimates with neighbors ($\rightleftarrows$). (a) Standard DFL yields a pilot estimate $\widehat\theta_{\rm init}$, which the large gradients of abnormal clients bias away from $\theta_0$. (b) At $\widehat\theta_{\rm init}$, normal local losses are flat and abnormal ones steep, so the trust weight $\widehat\gamma_m$ in \eqref{eq:adaptive_weight} is near one for normal and near zero for abnormal clients. (c) Rerunning DFL with local steps scaled by $\widehat\gamma_m$ turns abnormal clients into relays, and all estimates concentrate around $\theta_0$.}
\label{fig:overview}
\end{figure}

To solve those aforementioned problems, we propose here an adaptive decentralized federated learning (aDFL) approach for robust estimation. The key idea of aDFL is to dynamically adjust the learning rates of individual clients based on their behavior in the DFL process. Intuitively, those clients with suspicious behavior in their estimated gradients should be given smaller learning rates. In contrast, larger learning rates should be given to those clients who behave more normally in their gradients. The consequence is that the negative effect of those abnormal clients can be well controlled and minimized in a fully automatic way{\blue; see Figure~\ref{fig:overview} for an illustration}. {\blue  Specifically, the suspiciousness of a client is measured by the norm of its local gradient evaluated at a consistent initial estimator. A normal client should have a small gradient norm at such an estimator, whereas an abnormal client tends to have a large one. The learning rate of each client is then rescaled by a decreasing function of this gradient norm, and the associated tuning parameter is selected by a decentralized cross-validation procedure. We are able to show theoretically that, under appropriate regularity conditions, the resulting aDFL estimator achieves the same asymptotic efficiency as the estimator computed by normal clients only. Extensive numerical studies on both synthetic and real data further demonstrate that aDFL outperforms several competing methods across various corruption types and network structures.} {\blue Compared with existing approaches, our theory imposes no stringent requirements on the number of normal neighbors or the connectivity of the normal-client subgraph, and the algorithm requires no prior knowledge of client reliability or the number of abnormal neighbors.}

In summary, we make the following contributions in this work. {\blue  \textbf{Algorithmically}}, we develop here a novel aDFL approach for robust learning. This method adapts to diverse DFL settings and data scenarios. {\blue In particular, it allows heterogeneous data distributions across normal clients as long as they share a common target parameter. To select the tuning parameter in practice, we also develop a pessimistic decentralized cross-validation procedure.}
\textbf{Theoretically}, the convergence rate of the aDFL algorithm is rigorously analyzed. {\blue Our analysis also quantifies the effect of network imbalance without requiring the weighting matrix to be symmetric or doubly stochastic.} Our results show that
aDFL can achieve the oracle property (i.e., the same asymptotic efficiency as the estimator computed by normal clients only) under appropriate regularity conditions.  

{The rest of the paper is organized as follows. Section~\ref{sec:review} reviews related work. Section~\ref{sec:Stand} introduces the standard DFL framework and studies the properties of its estimator. Section~\ref{sec:robust} develops weighted DFL and the proposed aDFL method, establishes their theoretical properties, and presents a decentralized cross-validation procedure for selecting $\lambda_n$. Section~\ref{sec:exp} reports simulation studies and a real data application. Section~\ref{sec:con} concludes with a discussion of limitations and future directions.}

\section{Related Work}
\label{sec:review}
\textbf{Decentralized federated learning.}
{\blue Existing DFL methods can be broadly grouped into two categories. The first category consists of decentralized consensus optimization methods, which enforce agreement among neighboring clients to achieve global consensus; examples include exact first-order methods \citep{shi2015extra}, proximal-gradient methods \citep{li2019decentralized}, and gradient tracking \citep{song2022compressed}. The second category consists of decentralized gradient descent methods, which combine neighborhood averaging with local gradient updates \citep{yuan2016convergence,wu2023network}. Recent studies have further investigated optimal decentralized composite optimization \citep{ye2025optimal} and statistical inference for heterogeneous decentralized $M$-estimation \citep{gu2024statistical}, alongside related communication-efficient methods for heterogeneous distributed data \citep{fan2023communication,duan2022heterogeneity}; see \citet{beltran2023decentralized} for a comprehensive review. However, these methods assume that all participating clients are trustworthy and can suffer severe bias when some clients hold corrupted data.}

\textbf{Robust centralized learning.}
{\blue To handle abnormal clients, robust centralized learning has been studied extensively \citep{blanchard2017machine,yin2018byzantine,tu2021variance} and mainly follows two directions. The first direction designs robust aggregation rules, closely related to robust statistical estimation and robust personalized federated learning \citep{liu2025personalized}. A common strategy replaces the sample mean of local gradients or parameter estimates with a robust counterpart, such as the trimmed mean, coordinate-wise median \citep{yin2018byzantine}, or variance-reduced median-of-means \citep{tu2021variance}. The second direction first detects abnormal clients and then excludes them from subsequent aggregation, connecting naturally to outlier detection in statistics. For instance, \citet{qian2024bymi} combine sample splitting with false discovery rate (FDR) control to identify abnormal clients, while \citet{zhang2026mixture} compare local mixture densities to filter out suspicious estimates before mixture reduction in finite mixture models. Because these approaches rely on a central server to access all client updates simultaneously, they cannot be applied directly to decentralized networks.}

\textbf{Robust decentralized learning.}
{\blue Without a central server, robust DFL methods typically adapt the above two strategies to local neighborhoods. Along the robust aggregation direction, existing approaches employ clipping and trimming \citep{fang2022bridge,he2022selfcentered}, iterative outlier filtering \citep{wu2023byzantine}, and robust aggregation with gradient tracking and proximal updates for high-dimensional $M$-estimation \citep{zhang2024119808high}; related robust-loss methods include decentralized surrogate median regression for sparse estimation \citep{liu2022fast}. Along the detection direction, \citet{zhang2026exact} use sample-splitting score statistics with FDR control to identify and remove abnormal clients from the network.}

{\blue  In summary, existing methods have the following limitations in the presence of abnormal clients. Standard DFL methods are not robust to corrupted data, and robust centralized methods rely on a central server that is unavailable in DFL. Although robust DFL methods eliminate this reliance, they operate only on local neighborhoods. Consequently, neighborhood-based robust aggregation often requires a sufficient number of normal neighbors or a known upper bound on the number of abnormal neighbors, whereas hard removal requires the subgraph of normal clients to remain connected. Such requirements can be difficult to verify in sparse networks where each client has only a few neighbors. This motivates us to develop the aDFL method, which preserves the original communication network and continuously downweights suspicious clients through adaptive learning rates. As a result, aDFL avoids both local neighbor-count restrictions and subgraph connectivity requirements.}

\section{Standard Decentralized Federated Learning}
\label{sec:Stand}

We begin by introducing the model setup and notation.  Let $I_p$ be the $p \times p$ identity matrix. Define $\mathbf{1}_M = (1,\dots,1)^\top \in \mathbb{R}^M$ and $I^* = \mathbf{1}_M \otimes I_p \in \mathbb{R}^{Mp \times p}$. For a sequence $\{a^{(t)}\}$, define $a^{(\infty)} = \lim_{t \to \infty} a^{(t)}$. For two positive sequences $\{a_n\}$ and $\{b_n\}$, write $a_n \ll b_n$ or $a_n=o(b_n)$ if $a_n / b_n \to 0$ as $n \to \infty$. Write \(a_n \lesssim b_n\) or $a_n = O(b_n)$ if $a_n / b_n \leq C < \infty$ as $n \to \infty.$ 
For a vector $x \in \mathbb{R}^p$, denote its Euclidean norm by $\|x\|$. For a symmetric matrix $B \in \mathbb{R}^{p \times p}$, denote its smallest and largest eigenvalues by $\lambda_{\min}(B)$ and $\lambda_{\max}(B)$, respectively. For an arbitrary matrix $B \in \mathbb{R}^{p_1 \times p_2}$, define its $\ell_2$-norm as $\|B\| = \lambda^{1/2}_{\max}(B^\top B)$. For a set $S$, denote its cardinality by $|S|$  and represent its complement by $S^c$. Let $\Mean_m(\cdot)$ denote expectation with respect to the probability distribution $\mP_m$. The generic absolute constants $c$ and $C$ may vary from line to line. 

\subsection{Problem Description}
Assume a total of $N$ instances denoted as $(X_i,Y_i)$ for $1 \leq i \leq N.$ Here, $X_i =(X_{ij}) \in \mR^p$ is the feature vector and $Y_i \in \mR$ is the associated univariate response. 
We next consider a total of $M$ clients indexed by $\mathcal{M} = \{1,2,\dots, M\}.$ Let $\mathcal{S}_F = \{1,2,\dots,N\}$ represent the whole sample set, and let $\mathcal{S}_m$ denote the sample collected by the $m$th client. We then should have $\mathcal{S}_F = \bigcup_m \mS_m$ and $\mS_{m_1} \bigcap \mS_{m_2} = \emptyset$ for any $m_1 \neq m_2.$ 
For simplicity, we assume that $|\mS_m| = N/M = n$ for every $1 \leq m \leq M.$ In federated learning, data across different clients often exhibit considerable heterogeneity, which may arise from varied data collection environments. Despite the heterogeneity here,
we assume that all normal clients share a common underlying regression relationship. To be more precise, denote the joint distribution of $(x,y) \in \mathcal{X} \times \mathcal{Y}$ by $\mathcal{P}(x,y)$. Then we allow the marginal distributions $\mathcal{P}(x)$ and $\mathcal{P}(y)$ to be heterogeneous but require the conditional distribution $\mathcal{P}(y\mid x)$ to be the same across normal clients. The common parameter of interest is denoted as $\theta_0 \in \mR^p.$ Next, let $\ell(x,y;\theta)$ be a loss function with 
parameter $\theta \in \mR^p.$ Define a global loss function as 
$
\mL(\theta) = N^{-1} \sum_{i=1}^N \ell(X_i,Y_i;\theta).
$
It can be decomposed as 
$
\mL(\theta) = M^{-1} \sum_{m=1}^M \mL_m(\theta),
$
where 
$
\mL_m(\theta) = n^{-1} \sum_{i \in \mS_m} \ell(X_i,Y_i;\theta)
$
is the loss function defined on the $m$th client. { Let $\boldsymbol{\Theta}\subset\mR^p$ be the parameter space in Condition~\ref{ass:ps}.} Next define 
$
{ \wh \theta = \arg \min_{\theta\in\boldsymbol{\Theta}} \mL(\theta)}
$
as the whole sample estimator and 
$
{ \wh \theta_{m} = \arg \min_{\theta\in\boldsymbol{\Theta}} \mL_m(\theta)}
$
as the local estimator computed on the $m$th client. 
{ The population minimizer on client $m$ is $\theta_m=\arg\min_{\theta\in\boldsymbol{\Theta}}\Mean_m\ell(X_i,Y_i;\theta)$, with $\theta_m=\theta_0$ for normal clients.}

In this work, we consider the \textit{data-contaminated adversary} setting, where all of the clients are assumed to follow the learning protocol but the local data on the abnormal client may be corrupted \citep{jagielski2018manipulating}.  Specifically, define for each client a binary variable $a_m \in \{1,0\}$ to indicate whether the $m$th client is abnormal or not. Collect the indices of abnormal clients by $\mathcal{A} = \{m : a_m = 1\}$. Let $\varrho = |\mathcal{A}|/M \in [0, 1/2)$ be the fraction of  abnormal clients. Accordingly,  we assume that as $n \to \infty$,
$$
\begin{cases} 
\sqrt{n}(\wh \theta_m - \theta_0) \to_d N(0,\Sigma_m) & \text{if } m\notin \mathcal{A}, \\
\sqrt{n}(\wh \theta_m - \theta_m) \to_d N(0,\Sigma_m) \text{ with } \theta_m \neq \theta_0 &  \text{ if } m\in  \mathcal{A}.
\end{cases}
$$
for some positive definite matrix $\Sigma_m \in \mR^{p \times p}$.
Since $\theta_m \neq \theta_0$ for any $m \in \mA$, including those abnormal clients $\mA$ in DFL without effective control should cause seriously biased results. 

\subsection{The DFL Framework}
\label{sec:DFLfw}
We start with a standard DFL framework involving two key steps \citep{yuan2016convergence,wu2023network}. 
First, each client aggregates information from its neighbors to derive a neighborhood-averaged parameter estimator. Next, it updates this estimator by the method of gradient descent based on the data placed on the local client. Specifically, assume $M$ clients are connected through a communication network represented by an adjacency matrix $A = (a_{m_1 m_2}) \in \mathbb{R}^{M \times M}$. Here, $a_{m_1 m_2} = 1$ if client $m_1$ can receive information from client $m_2$, and $a_{m_1 m_2} = 0$ otherwise. Define in-degree $d_{m_1} = \sum_{m_2} a_{m_1 m_2}.$ We assume that $d_{m_1} > 0$ for every $1 \leq m_1 \leq M.$ 
Define the weighting matrix
$W = (w_{m_1 m_2}) \in \mathbb{R}^{M \times M}$ with $w_{m_1 m_2} = a_{m_1 m_2} / d_{m_1}$. Let $\wh \theta^{(t,m)}$ be the estimator obtained on the $m$th client at the $t$th iteration. Then, the update formula at the $(t+1)$th iteration is:
\beqr
\wt \theta^{(t,m)} = \sum_{k=1}^M w_{mk} \wh \theta^{(t,k)}; \quad \quad
\wh \theta^{(t+1,m)} = \wt \theta^{(t,m)} - \alpha \dot{\mL}_{m}\big(\wt \theta^{(t,m)}\big). \label{eq:ngd}
\eeqr
Here $\dot{\mathcal{L}}_{m}(\theta) \in \mR^p$ denotes the first order derivative of $\mathcal{L}_{m}(\cdot)$ with respect to $\theta$, and $\alpha \in \mR^{+}$ denotes the learning rate. Under appropriate regularity assumptions and assuming $\varrho=0$, \citet{wu2023network} showed that, with a sufficiently small $\alpha$ and a relatively balanced network structure $W$, $\wh \theta^{(t,m)}$ should converge numerically to an asymptotically efficient estimator of $\theta_0$. 
However, it is unclear what would happen if some of the clients are abnormal (i.e., $\varrho>0$). We are thus inspired to study the theoretical properties of $\wh \theta^{(t,m)}$ under the assumption with $\varrho>0$. {\blue  By \citet{wu2023network}, the balanceness of a network can be defined as} ${ \operatorname{SE}}^2 (W) = M^{-1}\| W^\top \textbf{1}_M - \textbf{1}_M \|^2$. 
In the most ideal situation with doubly stochastic $W$ in the sense that $\textbf{1}_M^\top W = \textbf{1}_M^\top$ \citep{li2019decentralized}, we have \(\SEW = 0\). Then we have the following regularity conditions.
\begin{assumption}[Parameter space]\label{ass:ps} Assume the parameter space $\boldsymbol{\Theta}$ is a compact and convex subset of $\mathbb{R}^p$. Let $\operatorname{int}(\boldsymbol{\Theta})$ be the set of interior points of $\boldsymbol{\Theta}$. Assume $\theta_m \in \operatorname{int}(\boldsymbol{\Theta})$ for $m \in \mA \cup \{ 0 \}$. Moreover, define $r = \sup_{\theta \in \boldsymbol{\Theta}} \max_m \| \theta - \theta_m \| > 0$ as a rough measure for the radius of $\boldsymbol{\Theta}$.
\end{assumption}
\begin{assumption}[Covariates distribution]\label{ass:distribution} { The observations $\{(X_i,Y_i):i\in\mathcal S_F\}$ are mutually independent, and those on client $m$ are identically distributed according to $\mP_m$.} 
\end{assumption}
\begin{assumption}[Local strong convexity]\label{ass:local:convexity}  Define $\Omega_m(\theta)=\Mean_m\big[ \ddot{\ell}(X_i,Y_i;\theta)\big]$ for $m \in \mM$, where $\ddot{\ell}(x,y;\theta) \in \mR^{p \times p}$ denotes the second order derivative of $\ell(x,y;\theta)$ with respect to $\theta$.
Assume that for $m \in \mM$, we have $\lambda_{\min}\big\{\Omega_m(\theta_m) \big\} \geq \lambda_{\min}$ for some positive constant $\lambda_{\min}$, and 
$\min_{m}\inf_{\theta \in \boldsymbol{\Theta}} \lambda_{\min}\big\{\ddot{\mL}_m(\theta) \big\} \geq 0$.
\end{assumption}
\renewcommand{\theassumption}{\arabic{assumption}} 
\begin{assumption}[Smoothness]
\label{ass:smoothness}Assume that
there exists some constant $C_{\max}>0$ such that for  $m \in \mM$,
\begin{gather*}
\sup_{\theta \in \boldsymbol{\Theta}} \Mean_m\big\{ \big\| \dot{\ell}(X_i,Y_i;\theta) - \Mean_m \big\{ \dot{\ell}(X_i,Y_i;\theta) \big\} \big\|_2^8 \big\}  \leq  C_{\max}^8,\\
\text{and } \sup_{\theta \in \boldsymbol{\Theta}}\Mean_m\big\{\big\| \ddot{\ell}(X_i,Y_i;\theta) - \Omega_m(\theta) \big\|_2^8 \big\}  \leq  C_{\max}^8.
\end{gather*}
Moreover, for any $i \in \mS_m$, $\dot{\ell}(X_i,Y_i;\theta) $ and $\ddot{\ell}(X_i,Y_i;\theta) $ are Lipschitz continuous in the sense that for any $\theta^\prime, \theta^{\prime \prime} \in \boldsymbol{\Theta}$, the following inequality holds
\begin{gather*}
\big\|
\dot{\ell}(X_i,Y_i;\theta^{\prime}) - \dot{\ell}(X_i,Y_i;\theta^{\prime\prime} ) \big\| \leq L(X_i,Y_i) \big\| \theta^{\prime} - \theta^{\prime\prime} \big\|,\\
\big\|\ddot{\ell}(X_i,Y_i;\theta^{\prime}) - \ddot{\ell}(X_i,Y_i;\theta^{\prime\prime} ) \big\| \leq L(X_i,Y_i) \big\| \theta^{\prime} - \theta^{\prime\prime} \big\| \nonumber
\end{gather*}
for some positive function $L(X_i,Y_i)$ and constant $L_{\max}$ such that 
$\Mean_m\big\{ L^8(X_i,Y_i) \big\} \leq L_{\max}^8.$ 
\end{assumption}
\begin{assumption}[Network structure] 
\label{ass:network} 
There exists some constant $\rho \in (0,1)$ such that 
$\|W^\top (I_M-M^{-1}\mathbf{1}_M\mathbf{1}_M^\top) W\|+\operatorname{SE}^2(W)\leq\rho^2.$
\end{assumption}
\begin{assumption} [Non-vanishing bias] 
\label{ass:bias}
Define 
\(\flat_m = \mathbb{E}_m\{ \dot{\ell}(X_i,Y_i;\theta_0)  \}\),  
Assume 
$\min_{m \in \mA} \|\flat_m\| \geq \flat_{\min}$ for some constant  $\flat_{\min} > 0$ { when $\mA\ne\varnothing$.} 
\end{assumption}

Condition \ref{ass:ps} defines the parameter space for $\theta_m$ with $m \in \mA \cup \{ 0 \}$. 
Similar conditions are also used in \citet{zhang2013communication} and \citet{jordan2019communication}. 
Condition \ref{ass:distribution} addresses the distribution of the data $\{(X_i,Y_i): i \in \mS_k\}$, allowing the data distributions to vary across different clients. This relaxes the homogeneous data condition commonly assumed in existing approaches \citep{fang2022bridge, qian2024bymi}. Condition \ref{ass:local:convexity} { assumes convex losses with local strong convexity near the population minimizers; it does not require global strong convexity. Related localization arguments are used in \citet{zhang2013communication}.}
This makes our theoretical results applicable to a broader class of loss functions. 
Condition \ref{ass:smoothness} requires the local loss functions to be sufficiently smooth, which is a classical regularity condition in convex optimization \citep{jordan2019communication} and federated learning \citep{zhang2024119808high}.  Condition \ref{ass:network} is a condition about the network structure. 
{ The condition permits imbalanced row-stochastic matrices, relaxing the balance requirement of the doubly stochastic assumption used in the literature \citep{li2019decentralized,gu2024statistical}.} Condition \ref{ass:bias} forces abnormal clients to be distinguishable from normal clients since $\|\flat_m\| = 0$ for any $m \notin \mA$. 

We start with the properties of the whole-sample estimator $\widehat{\theta}$ with $\varrho > 0.$ {\blue Let $\theta_{\mA}=\arg\min_{\theta\in\boldsymbol{\Theta}}\Mean\mL(\theta)$ be the pooled population risk minimizer, and define
$
V(\wh\theta)=\Mean\|\wh\theta-\theta_{\mA}\|^2$ and $
B(\wh\theta)=\|\theta_{\mA}-\theta_0\|^2+
2(\theta_{\mA}-\theta_0)^\top\Mean(\wh\theta-\theta_{\mA}).
$
Let $\Omega_{\mA}=\Omega_{\mA}(\theta_{\mA})$ and $\Sigma_{\mA}=\Sigma_{\mA}(\theta_{\mA})$.} This leads to the following Theorem \ref{thm:ws_estimator} about the mean-squared error (MSE) of ${\widehat{\theta}}$.

\begin{theorem}[MSE of $\wh \theta$]
\label{thm:ws_estimator} 
Assume Conditions \ref{ass:ps} -- \ref{ass:bias} hold. { Further assume that $\varrho<\epsilon${ , where $\epsilon>0$ is a sufficiently small constant.} Then
$\Mean\|\wh\theta-\theta_0\|^2=V(\wh\theta)+B(\wh\theta)$, where
$V(\wh\theta)\lesssim C_{\max}^2/\{(1-\varrho)^2\lambda_{\min}^2N\}+O(N^{-2})$ and
$NV(\wh\theta)-\operatorname{tr}\{\Omega_{\mA}^{-1}\Sigma_{\mA}\Omega_{\mA}^{-1}\}\to0$ as $N\to\infty$. Moreover,}
\[
{ \frac{\varrho^2\|\overline{\flat}_{\mA}\|^2}{L^2_{\max}}
- C\frac{\varrho\|\overline{\flat}_{\mA}\|}{N}
\le B(\wh\theta)
\le
\frac{4\varrho^2\|\overline{\flat}_{\mA}\|^2}{(1-\varrho)^2\lambda_{\min}^2}
+C\frac{\varrho\|\overline{\flat}_{\mA}\|}{N}.}
\]%
Here  $\overline{\flat}_{\mA} = |\mA|^{-1} \sum_{m \in \mA} \flat_m$ (defined as zero when $\mA$ is empty), 
\[
\Sigma_{\mA}(\theta)
=M^{-1}\sum_{m=1}^M\Mean_m\!\big[
\big\{\dot\ell_i(\theta)-\Mean_m\dot\ell_i(\theta)\big\}
\big\{\dot\ell_i(\theta)-\Mean_m\dot\ell_i(\theta)\big\}^{\top}
\big]
\]
and $\Omega_{\mA}(\theta)=M^{-1}\sum_{m=1}^M\Omega_m(\theta).$
\end{theorem}
By Theorem \ref{thm:ws_estimator}, we know that the MSE of $\wh \theta$ is mainly determined by two terms. The first term \(V(\wh\theta)\)  {\blue measures the mean-squared error about $\theta_{\mA}$}, with its leading term given by \(N^{-1} \operatorname{tr}\big\{ \Omega^{-1}_{\mA} \Sigma_{\mA} \Omega^{-1}_{\mA}\big\}\). The second term 
{ reflects the bias due to abnormal clients, with magnitude of order at most $\varrho^2\|\overline{\flat}_{\mA}\|^2+\varrho\|\overline{\flat}_{\mA}\|/N$.} If \(  \varrho \to 0\), \(\wh \theta\) remains to be a consistent estimator for \(\theta_0\). 
{ However, to make the contamination bias negligible at the root-$N$ scale, a sufficient condition is $\varrho^2=o(N^{-1})$. This leads to $n/M=o(|\mA|^{-2})$.}
Otherwise, \(\wh \theta\) may exhibit a non-negligible bias. 
However, this condition is not always achievable in practice. 
Consider for example a situation with each client representing a local hospital. In this case, each client (e.g., a hospital) might hold a sufficiently large amount of data. Nevertheless, the total number of clients (hospitals) is typically quite limited. According to classical results on DFL, under suitable assumptions, the difference between the standard DFL estimator and $\wh \theta$ is statistically ignorable. 

\subsection{The \texorpdfstring{{\blue  Numerical}}{Numerical} Properties of the Standard DFL Estimator}
\label{eq:standard:DFL}
Let $\wh \theta^{*(t)} = \{(\wh \theta^{(t,1)})^\top,\dots,(\wh \theta^{(t,M)})^\top \}^\top \in \mathbb{R}^{Mp}$ denote the stacked standard DFL estimator obtained at the \(t\)th iteration, and let $\wh \delta_0 = \max_m \|\wh \theta^{(0,m)} - \wh \theta\|$ 
be the initial distance. 
The numerical convergence property of the standard DFL algorithm is elaborated by the following Proposition \ref{prop:bad_estimator}. 
\begin{proposition}[Convergence Property of the Standard DFL]
\label{prop:bad_estimator}  { Assume the conditions of Theorem~\ref{thm:ws_estimator}. Let $\bar\theta^{(0)}=M^{-1}\sum_m\widehat\theta^{(0,m)}$. Further assume that $\alpha\sqrt M+\SEW<\epsilon$ and $\|\bar\theta^{(0)}-\widehat\theta\|+\|\widehat\theta^{*(0)}-I^*\bar\theta^{(0)}\|<\epsilon$, where { $\epsilon>0$ is a sufficiently small constant.} }
Then, with probability at least $1 - O(M/n^4)$, the following relationship holds.
\[
M^{-1/2}\Big\|\widehat{\theta}^{*(t)} - I^*\wh \theta\Big\|  \lesssim \Big( 1-
\frac{\alpha (1-\varrho)\lambda_{\min}}{8}\Big)^{t} \wh\delta_0 + \frac{\alpha + \SEW }{(1-\rho)(1-\varrho)\lambda_{\min}}  .
\]
\end{proposition}
\noindent
Proposition \ref{prop:bad_estimator} suggests that
the {normalized discrepancy} between the DFL estimator $\wh \theta^{*(t)}$ obtained in the $t$th step and the whole-sample estimator $I^* \wh \theta$ is {bounded, up to a constant, by the sum of} (1) the optimization error ${\{1 - \alpha(1-\varrho)\lambda_{\min}/8\}^{t}\wh\delta_0}$ and (2) the {network and step-size error} $\big\{\alpha + \SEW\big\} / {\big\{(1-\rho)(1-\varrho)\lambda_{\min}\big\}}$. {As $t\to\infty$,} we obtain $M^{-1/2} \| \wh \theta^{*(\infty)} - I^* \wh \theta \| {\lesssim} \big\{\alpha + \SEW\big\}/ {\big\{(1-\rho)(1-\varrho)\lambda_{\min}\big\}}$. {Therefore, $\alpha+\SEW=o(N^{-1/2})$ suffices to make this normalized discrepancy statistically negligible at the root-$N$ scale.} Moreover, we can combine the conclusions of Theorem \ref{thm:ws_estimator} and Proposition \ref{prop:bad_estimator} to obtain an explicit bound on $\|\wh \theta^{*(\infty)} - I^* \theta_0\|$ in terms of $n, M$, and $\rho$ as
$$
M^{-1/2}\Big\|\wh \theta^{*(\infty)} - I^* \theta_0\Big\|  \lesssim \frac{\alpha + \operatorname{SE}(W) }{(1-\rho)(1-\varrho)} + \frac{1}{\sqrt{\delta}} \Big\{ \frac{1}{(1 - \varrho)\sqrt{nM}} + \varrho \|\overline{\flat}_{\mathcal{A}} \| + \frac{(\varrho\|\overline{\flat}_{\mathcal{A}}\|)^{1/2}}{\sqrt{N}} + \frac{1}{nM}\Big\}  \nonumber
$$
with probability at least { $1-\delta-O(M/n^4)$} for some small constant $\delta>0$.

However, Theorem \ref{thm:ws_estimator} indicates that \(\wh \theta\) itself might be biased. 
Consequently, the DFL estimator is also expected to suffer from the same bias.
This motivates us to develop a robust DFL algorithm, so that the negative effects due to the abnormal clients can be well controlled.

\section{Robust DFL}
\label{sec:robust}
\subsection{Weighted Decentralized Federated Learning}

By Theorem \ref{thm:ws_estimator}, we know that the key reason responsible for the poor performance of the standard DFL estimator 
is the existence of the abnormal clients (i.e., $\mA$).  Unfortunately, a standard DFL algorithm treats those abnormal clients and normal clients equally without differentiating their relative trustworthiness. One natural solution is to revise \( \alpha \) slightly so that different learning rates can be used for different clients according to their trustworthiness. Intuitively, larger learning rates should be given
to clients, which are more likely to have \( a_m = 0 \). In contrast, significantly reduced learning rates should be given to those which are more likely to have $a_m=1$.  
Accordingly, the bias due to those abnormal clients in \( \mathcal{A} \) can be greatly reduced. {\blue  To this end, let} $\wh \theta^{(t,m)}_{\mA}$ be an estimator obtained on the $m$th client at the $t$th iteration. We are motivated to modify the standard DFL updating formula \eqref{eq:ngd} as:
\beqr
\label{eq:bt:ngd}
\wh \theta^{(t+1,m)}_{\mA} &=& \wt \theta^{(t,m)}_{\mA} - \alpha  \gamma_m \dot{\mL}_{m}\big(\wt \theta^{(t,m)}_{\mA}\big)
\eeqr
with $\wt \theta^{(t,m)}_{\mA} = \sum_{k} w_{mk}\wh \theta_{\mA}^{(t,k)}.$ 
Here, \(\gamma_m \in [0,1]\) is a data-driven weight that reflects the trustworthiness of the \(m\)th client. 

Subsequently, we analyze the theoretical properties of the algorithm \eqref{eq:bt:ngd} with general \(\gamma_m\)s. 
In particular, we are eager to understand the role played by the adaptive weights $\gamma_m$s. 
To this end, define 
$
\bar{\Delta}^2_2 = M^{-1} \sum_{m=1}^M \big\{\gamma_m - (1-a_m) \big\}^2
$
as the mean squared distance between $\gamma_m$ and the oracle weight $1-a_m$. Write
$\bar{\gamma}^{\mg} = M^{-1} \sum_{m \notin \mA} \gamma_m$
, $\omegaavg = ( |\mA|^{-1} \sum_{m \in \mA} \gamma_m^2 )^{1/2}$ and $\flatavg = ( |\mA|^{-1} \sum_{m \in \mA} \|\flat_m\|^2 )^{1/2}.$ Further denote \(\wh \theta^{*(t)}_{\mA} = \{ ( \wh \theta_{\mA}^{(t,1)}  )^\top,\dots,( \wh \theta_{\mA}^{(t,M)})^\top \}^\top \in \mathbb{R}^{Mp}\) as the stacked estimator obtained from Equation \eqref{eq:bt:ngd} at the $t$th iteration. 
For theoretical purposes, define an oracle estimator as the estimator obtained by using data from the trustworthy clients only. Denote this oracle estimator by 
$
\wh \theta_{\mA} =  \argmin_{\theta\in\boldsymbol\Theta}  \sum_{m \notin \mA}  \mL_m(\theta).
$ 
Let $\wh \delta_0^{\mA} = \max_m \|\wh \theta^{(0,m)}_{\mA} - \wh \theta_{\mA}\|$.
We then have the following Theorem \ref{thm:robust_estimator}.

\begin{theorem}[Convergence property of $\wh \theta^{*(t)}_{\mA}$]
\label{thm:robust_estimator}
Assume Conditions~\ref{ass:ps}--\ref{ass:bias} hold. For deterministic weights $\{\gamma_m\}_{m=1}^M\subset[0,1]$, assume that $\bar\gamma^{\mg}$ is bounded below by a fixed positive constant. Let $\bar\theta_{\mA}^{(0)}=M^{-1}\sum_m\widehat\theta_{\mA}^{(0,m)}$. Further assume that $\alpha\sqrt M+\SEW<\epsilon$, $\varrho\omegaavg<\epsilon$, and
$\|\bar\theta_{\mA}^{(0)}-\widehat\theta_{\mA}\|+\|\widehat\theta_{\mA}^{*(0)}-I^*\bar\theta_{\mA}^{(0)}\|\leq\epsilon$.
{ Here $\epsilon>0$ is a sufficiently small constant.} 
Then, with probability at least $1-O\{M/n^4+1/(\log n)^4\}$, we have
$M^{-1/2}
\|\widehat{\theta}_{\mA}^{*(t)} - I^*\,\widehat{\theta}_{\mA}\|
\lesssim \operatorname{Err}_1 + \operatorname{Err}_2 + \operatorname{Err}_3,
$
where 
\beqrs
\operatorname{Err}_1 &=& \Bigl(1 - \frac{\alpha\bar{\gamma}^{\mg} \lambda_{\min}}{8}\Bigr)^{t} \widehat{\delta}_0^{\mA}, \; \\
\operatorname{Err}_2 &=& \frac{\alpha L_{\max} + \SEW}{(1-\rho) \lambda_{\min} \bar{\gamma}^{\mg}}
 \Bigl\{\Bigl(\frac{\log n}{n}\Bigr)^{1/2} L_{\max} 
+ \varrho^{1/2} \flatavg\Bigr\}, 
\\
\operatorname{Err}_3 &=& \frac{1}{\bar{\gamma}^{\mg} \lambda_{\min}}
\biggl[\varrho \flatavg \omegaavg
\!+\!\bar{\Delta}_2 \Bigl\{\Bigl(\frac{\log N}{N}\Bigr)^{\!\!\frac12} 
+ L_{\max} \|\widehat{\theta}_{\mA} - \theta_0\|\Bigr\} \biggr].
\eeqrs
{ If $M=o(n^4)$ as $n\to\infty$, the iterates converge with probability tending to one, and
\begin{gather}
\label{eq:rngd:asymptotic}
M^{-1/2}\|\widehat\theta_{\mA}^{*(\infty)}-I^*\widehat\theta_{\mA}\|
=O_p\left[\frac{1}{\bar\gamma^{\mg}}
\left\{(\alpha+\SEW)(n^{-1/2}+\varrho^{1/2})
+\varrho\omegaavg+\frac{\bar\Delta_2}{\sqrt N}\right\}\right].
\end{gather}
}
\end{theorem}
\noindent
Compared to the classical results on DFL (see Proposition \ref{prop:bad_estimator} for details), the
main difference of Theorem \ref{thm:robust_estimator} is the inclusion of an additional statistical error term $\operatorname{Err}_3$. 
If oracle weights $(1 - a_m)$s are employed, we then have $\bar{\gamma}^{\mg}=1-\varrho$ and \(\bar{\Delta}_2 = \omegaavg \equiv 0\). Accordingly, the influence of abnormal clients on $\wh \theta^{*(t)}_{\mA}$ can be eliminated completely, as long as the learning rate $\alpha$ is sufficiently small and the network structure $W$ is sufficiently balanced. 

For \({ \wh \theta_{\mA}^{*(\infty)}}\) to achieve the oracle property, the right-hand side of Equation \eqref{eq:rngd:asymptotic} should be of an $o_p(1/\sqrt{N})$ order. This conclusion holds if the following three conditions can be satisfied. They are, respectively, 
 (1) \(\big\{\alpha + \SEW\big\}\big\{1/\sqrt{n} + \varrho^{1/2}\big\} / \bar{\gamma}^{\mg} = o(1/\sqrt{N})\), (2) \(\omegaavg / \bar{\gamma}^{\mg} =  o_p\big(1/(\varrho\sqrt{N})\big)\), and (3) $\bar{\Delta}_2/\bar{\gamma}^{\mg} = o_p(1)$. The first condition can be satisfied by setting a reasonable $\bar{\gamma}^{\mg}$, and a sufficiently small $\alpha$ and \(\SEW\). Both conditions (2) and (3) require $\gamma_m$ to approximate the oracle weights $(1 - a_m)$ closely. 
 However, since the status of the clients is unknown in advance, we need to develop an effective estimator for $\gamma_m$ so that both conditions (2) and (3) can be practically satisfied.

\subsection{Adaptive Decentralized Federated Learning}
\label{sec:adfl}
To this end, an effective measure for the trustworthiness of a client is necessarily needed. Note that a trustworthy client should have a small gradient norm at a reasonably accurate parameter estimator.
In contrast, an abnormal client tends to exhibit a larger gradient norm. Thus, the size of the gradient norm might serve as a natural indicator of trustworthiness. Based on this idea, we develop below a two-stage algorithm.

{\sc Stage 1}. We start with assuming for each client $m$ an initial 
estimator, denoted by \(\wh \theta_{\operatorname{init}}^{(m)}\),  which may not be statistically efficient but must be consistent. For example, one might use the standard DFL estimator as described in Section \ref{sec:DFLfw} to serve this purpose, if condition \( \varrho \to 0\) can be well satisfied. 

{\sc Stage 2}. Once the initial estimator $\wh \theta_{\operatorname{init}}^{(m)}$ is obtained, the adaptive weight for the $m$th client can be computed as
\beq
\label{eq:adaptive_weight}
\wh \gamma_m = \pi\big\{ \lambda_n \big\|\dot{\mL}_{m}(\wh \theta_{\operatorname{init}}^{(m)}) \big\| \big\},
\eeq
where $\pi(\cdot) \in [0, 1]$ is an appropriately selected and monotonically decreasing mapping function. For example, we use $\pi(x) = \exp(-x)$ in this work. Moreover, $\lambda_n$ is a positive tuning parameter, which controls the gradient scale. It is important to note that the selection of $\lambda_n$ plays a critical role in this algorithm. Specifically, $\lambda_n \|\dot{\mathcal{L}}_{m}(\wh \theta_{\operatorname{init}}^{(m)})\|$ should not be too low. Otherwise, \(\wh{\gamma}_m\) cannot shrink to $0$ quickly 
for those abnormal clients. Conversely, this product should not be too large either. Otherwise, \(\wh{\gamma}_m\)  might not give sufficient trust to those trustworthy clients.  
Subsequently, the updating step in Equation \eqref{eq:bt:ngd} can be executed by replacing  $\gamma_m$  with \(\wh \gamma_m\) in \eqref{eq:adaptive_weight}. This leads to a practically feasible aDFL estimator \(\wh \theta^{(t,m)}_{\operatorname{aDFL}}\) for the $m$th client at the $t$th iteration with \(\wh \theta^{(0,m)}_{\operatorname{aDFL}} = \wh \theta_{\operatorname{init}}^{(m)}\).  The pseudo code for the aDFL algorithm is described below in Algorithm \ref{alg:multi_stage:aDFL}. 
\begin{algorithm}[htb]
   \caption{Adaptive Decentralized Federated Learning}
\label{alg:multi_stage:aDFL}
   \begin{algorithmic}[1]
   \REQUIRE Network $W$, learning rate $\alpha$, { tuning parameter $\lambda_n$,} maximum number of iterations $T$. \\
   \ENSURE aDFL estimator $\{\wh \theta_{\text{aDFL}}^{(T,m)} \big\}_{m=1}^M.$
      \STATE Compute initial estimators $\{\wh \theta_{\operatorname{init}}^{(m)}\}_{m=1}^M$, and set $\wh \theta_{\text{aDFL}}^{(0,m)} = \wh \theta_{\operatorname{init}}^{(m)}$ for $1 \leq m \leq M.$
      \FOR{$0 \leq t \leq T-1$}
      \FOR{$1 \leq m \leq M$ (distributedly)}
      \STATE Compute the neighborhood-averaged estimator $\wt \theta^{(t,m)}_{\text{aDFL}} = \sum_k w_{mk} \wh \theta^{(t,k)}_{\text{aDFL}}.$
      \STATE Compute  
      $
      \wh \theta^{(t+1,m)}_{\text{aDFL}}= \wt \theta^{(t,m)}_{\text{aDFL}} - \alpha \wh \gamma_m \dot{\mL}_{m} (\wt \theta^{(t,m)}_{\text{aDFL}}),
      $ where $\wh \gamma_m$ is given by \eqref{eq:adaptive_weight}.
      \ENDFOR
       \ENDFOR
   \end{algorithmic}
\end{algorithm}

We next study the theoretical properties of the proposed aDFL estimator $\wh \theta^{(t,m)}_{\operatorname{aDFL}}$. 
Denote the stacked aDFL estimator at iteration $t$ as
$
\widehat{\theta}^{*(t)}_{\operatorname{aDFL}} = \big\{\, (\widehat{\theta}_{\operatorname{aDFL}}^{(t,1)})^\top, \dots, (\widehat{\theta}_{\operatorname{aDFL}}^{(t,M)})^\top \,\big\}^\top \in \mathbb{R}^{Mp}.
$
Write the corresponding estimators of $\omegaavg$, $\bar{\Delta}_2$,and $\bar{\gamma}^{\mg}$ based on Equation \eqref{eq:adaptive_weight} as $\hat{\bar{\gamma}}^{\mA}_2,\hat{\bar{\Delta}}_2$ and $\hat{\bar{\gamma}}^{\mg}$, respectively. 
We then have the following Theorem \ref{thm:two_stage:robust_estimator}. 
\begin{theorem}[Convergence rate of the aDFL]
\label{thm:two_stage:robust_estimator}
{ Assume the conditions of Theorem~\ref{thm:ws_estimator}. Let $\pi(x)=\exp(-x)$, and let $\widehat\theta_{\operatorname{init}}^{(m)}$ be the standard DFL estimator at numerical convergence, with the initialization specified in Proposition~\ref{prop:bad_estimator}. Assume $C_0\log N\leq\lambda_n\lesssim(n/\log n)^{1/2}$ and $\alpha+\SEW\lesssim N^{-1/2}$, where $C_0>2/\flat_{\min}$. }Then, with probability at least $1-O\{M/n^4+1/(\log n)^4+N^{-1}+\varrho^2\}$, we have: (1) $\varrho\hat{\bar{\gamma}}^{\mA}_2\lesssim N^{-1/2}$, (2) $\hat{\bar{\Delta}}_2^2\lesssim\lambda_n^2\{\log n/n+\|\wh\theta-\theta_0\|^2\}$, and (3) $1/\hat{\bar{\gamma}}^{\mg}\lesssim\exp(c\lambda_n\|\wh\theta-\theta_0\|)$. { Further assume $M\to\infty$, $M\lambda_n^2=o(n)$, $\varrho=o(1)$, $\lambda_n(\log n/n)^{1/2}=o(1)$, and $\lambda_n\|\widehat\theta-\theta_0\|=O_p(1)$. The adaptive iterates then converge with probability tending to one, and
\begin{equation}
\label{eq:rngd:asymptotic:stage2}
\begin{aligned}
&M^{-1/2}\|\widehat\theta_{\operatorname{aDFL}}^{*(\infty)}-I^*\widehat\theta_{\mA}\|=\exp\{C\lambda_n\|\widehat\theta-\theta_0\|\}
\left\{\frac{\lambda_n\|\widehat\theta-\theta_0\|}{\sqrt N}O_p(1)+o_p(N^{-1/2})\right\}.
\end{aligned}
\end{equation}
}
\end{theorem}

\noindent
From Theorem \ref{thm:two_stage:robust_estimator},  
the statistical error introduced by abnormal clients \eqref{eq:rngd:asymptotic:stage2} can be further reduced to be of the order $o_p(1 / \sqrt{N})$, if we can further assume that $\lambda_n \|\wh\theta - \theta_0 \| = o_p(1)$.  Here, recall that $\wh\theta$ denotes the whole-sample estimator. { The condition $M\lambda_n^2=o(n)$ controls the same-sample weighting remainder $\lambda_n/n$; for $\lambda_n\asymp\log N$, it becomes $M(\log N)^2=o(n)$.} 
This result implies that the aDFL estimator 
achieves the same asymptotic efficiency as  $\wh \theta_{\mA}$, as long as a suitable tuning parameter $\lambda_n$ can be used. 

{ The validity of Theorem~\ref{thm:two_stage:robust_estimator} relies on the assumption that an initial estimator of a reasonable quality be provided. It can be satisfied by a statistically consistent (but not necessarily efficient) initial estimator whose error satisfies the stated tuning-dependent rate. As shown by our Theorem~\ref{thm:ws_estimator} and Proposition~\ref{prop:bad_estimator}, a standard DFL estimator can serve the purpose with $\lambda_n\varrho=o(1)$.
In practice, one might also consider other decentralized robust estimators \citep{fang2022bridge,zhang2024119808high} as initial estimators $\widehat\theta_{\operatorname{init}}^{(m)}$ with $\varrho\in[0,1/2)$ under appropriate regularity conditions, including Condition~\ref{ass:consensus} below.}

\begin{assumption}[Consensus Convergence]
\label{ass:consensus}
Denote $\bar\theta_{\operatorname{init}}=M^{-1}\sum_{m=1}^M\widehat\theta_{\operatorname{init}}^{(m)}$. Assume that (1) $\lambda_n\|\bar\theta_{\operatorname{init}}-\theta_0\|=O_p(1)$; and (2)
$ 
(\lambda_n+\sqrt{N\varrho})
\{M^{-1}\sum_{m=1}^M\|\widehat\theta_{\operatorname{init}}^{(m)}-\bar\theta_{\operatorname{init}}\|^2\}^{1/2}=o_p(1).
$ 
{ When $\mA$ is empty, replace $\lambda_n$ in part~(2) by $\lambda_n\sqrt M$.}
\end{assumption}
Condition~\ref{ass:consensus} requires that the initial estimators $\{\widehat\theta_{\operatorname{init}}^{(m)}\}_m$ have a clear consensus in the sense that their root-mean-square disagreement satisfies the rate in part~(2). Under the preceding contamination-rate condition, such initial estimators can be obtained, for example, by (1) a standard DFL algorithm with a sufficiently small $\alpha+\operatorname{SE}(W)$ value; or (2) the EXTRA algorithm of \citet{shi2015extra}, with mixing matrices and a step size satisfying its convergence conditions.

We summarize this result in the following Corollary \ref{col:robust_estimator}.
\begin{corollary}[General initial estimator]
\label{col:robust_estimator}
Assume Conditions~\ref{ass:ps}--\ref{ass:bias} and \ref{ass:consensus}, and let $\pi(x)=\exp(-x)$. Further assume $M\to\infty$, $M\lambda_n^2=o(n)$, $\alpha+\SEW=O(N^{-1/2})$, and $C_0\log N\leq\lambda_n=o\{(n/\log n)^{1/2}\}$ for $C_0>2/\flat_{\min}$. The iterates converge with probability tending to one, and
\[
\begin{aligned}
&M^{-1/2}\|\widehat\theta_{\operatorname{aDFL}}^{*(\infty)}-I^*\widehat\theta_{\mA}\|=\exp\{C\lambda_n\|\bar\theta_{\operatorname{init}}-\theta_0\|\}
\left\{\frac{\lambda_n\|\bar\theta_{\operatorname{init}}-\theta_0\|}{\sqrt N}O_p(1)+o_p(N^{-1/2})\right\}.
\end{aligned}
\]
\end{corollary}
\noindent
{ We find that the aDFL estimator has the oracle property as long as $\lambda_n \|\bar{\theta}_{\operatorname{init}} - \theta_0\| = o_p(1)$.}

\begin{remark}[On the condition $\varrho=o(1)$]
\label{rem:varrho}
{The condition $\varrho=o(1)$ is used only to ensure that the standard DFL estimator is accurate enough to serve as the initial estimator, that is, $\lambda_n\|\bar\theta_{\operatorname{init}}-\theta_0\|=o_p(1)$; see Proposition~\ref{prop:bad_estimator} and Corollary~\ref{col:robust_estimator}. By Corollary~\ref{col:robust_estimator}, any initial estimator meeting this accuracy requirement suffices, including decentralized robust estimators that allow $\varrho\in[0,1/2)$. Intuitively, the weights only require the gradient norms of abnormal clients at the initial estimator to be well separated from those of normal clients. The multi-stage implementation in Algorithm~\ref{alg:extend:aDFL} reuses the updated estimates to recompute the trust weights. The aDFL estimator with the standard DFL initializer performs well in our experiments for abnormal fractions as large as $\varrho=0.3$; see Section~\ref{sec:exp}.}
\end{remark}

The numerical convergence speed and statistical efficiency of Algorithm \ref{alg:multi_stage:aDFL} can be improved in two ways. First, Theorem \ref{thm:two_stage:robust_estimator} indicates a convergence rate of { $1-O(\alpha\hat{\bar\gamma}^{\mg})$}. Thus, after computing $\widehat{\gamma}_m$, each client can obtain $\gamma_{\max} = \max_m\widehat{\gamma}_m$ { by max-consensus and then update} $\widehat{\gamma}_m \leftarrow \widehat{\gamma}_m/\gamma_{\max}$ so that $\hat{\bar{\gamma}}^{\mg}$ can be increased. Second, both Theorem \ref{thm:two_stage:robust_estimator} and Corollary \ref{col:robust_estimator} reveal that the error bound depends on $\|\bar{\theta}_{\operatorname{init}} - \theta_0\|$. Then the aDFL estimator can be used as a new initial estimator for Algorithm \ref{alg:multi_stage:aDFL} repeatedly. 
The multi-stage aDFL algorithm is provided in Algorithm \ref{alg:extend:aDFL}.

{

\makeatletter
\let\rdfl@makebox\float@makebox
\renewcommand{\float@makebox}[1]{\color@vbox\normalcolor\rdfl@makebox{#1}\color@endbox}
\makeatother
\begin{algorithm}[htb]
\caption{Multi-stage Adaptive Decentralized Federated Learning}
\label{alg:extend:aDFL}
\begin{algorithmic}[1]
\REQUIRE Initial estimators $\{\wh \theta^{(m)}_{\operatorname{init}}\}_{m=1}^M$; maximum number of iterations $T$; number of stages $S$; {network $W$, learning rate $\alpha$, and tuning parameter $\lambda_n$.}
\ENSURE aDFL estimators $\{\wh \theta_{\text{aDFL}}^{(T,m)} \big\}_{m=1}^M$
\FOR{$s=1,\ldots,S$}
\STATE Set $\wh \theta_{\text{aDFL}}^{(0,m)} = \wh \theta^{(m)}_{\operatorname{init}}$ for $1 \leq m \leq M$
\STATE {Recompute $\widehat\gamma_m$ from \eqref{eq:adaptive_weight} using the current $\widehat\theta_{\operatorname{init}}^{(m)}$, for $1\leq m\leq M$.}
\FOR{$0 \leq t \leq T-1$}
\FOR{$1 \leq m \leq M$ in parallel}
\STATE Compute the neighborhood-averaged estimator $\wt \theta^{(t,m)}_{\text{aDFL}} = \sum_k w_{mk} \wh \theta^{(t,k)}_{\text{aDFL}}$
\STATE Update 
\[
\wh \theta^{(t+1,m)}_{\text{aDFL}}= \ \wt \theta^{(t,m)}_{\text{aDFL}} - \alpha \wh \gamma_m \dot{\mL}_{m} (\wt \theta^{(t,m)}_{\text{aDFL}}).
\]
\ENDFOR
\ENDFOR
\STATE {Set $\widehat\theta_{\operatorname{init}}^{(m)}\leftarrow\widehat\theta_{\text{aDFL}}^{(T,m)}$ for $1\leq m\leq M$.}
\ENDFOR
\end{algorithmic}
\end{algorithm}}

\subsection{The Choice of \texorpdfstring{$\lambda_n$}{lambda n}}
\label{ap:subsec:cn}

{ Theorem~\ref{thm:two_stage:robust_estimator} and Corollary~\ref{col:robust_estimator} specify the asymptotic order of $\lambda_n$, but leave its finite-sample value unspecified. Since $\lambda_n$ directly governs the adaptive weights and the final estimator, choosing it in a data-driven manner is essential in practice. To this end, we propose here a decentralized cross-validation (DCV) method based on sample splitting.

Specifically, each client $m$ splits $\mathcal S_m$ into a training set $\mathcal S_m^{\rm tr}$ of size $n_{\rm tr}$ and an independent validation set $\mathcal S_m^{\rm val}$ of size $n_{\rm val}$. Let $\mathcal D_{\rm tr}$ denote the collection of all local training observations, which are used to compute the initial estimators and fit candidate models. For each $\lambda\in\Lambda_n=\{\lambda_n^1,\ldots,\lambda_n^K\}\subset[0,\infty)$, we run Algorithm~\ref{alg:multi_stage:aDFL} on $\mathcal D_{\rm tr}$ to obtain $\widehat\theta_m(\lambda)=\widehat\theta_{\rm aDFL}^{(T,m)}(\lambda)$ and $\widehat\gamma_m(\lambda)=\exp[-\lambda\|\dot{\mathcal L}^{\rm tr}_m\{\widehat\theta_{\rm init}^{(m)}\}\|]$, where $\mathcal L_m^{\rm tr}$ is the local training loss. Given a validation score $\ell_{\rm val}$, our goal is to select $\lambda$ that minimizes the average population validation risk over the normal clients $\mg=\mA^c$. With the oracle weights $\gamma_m=1-a_m$, this target risk is
\[
\mathcal L_{\mg}^{\rm val}(\lambda)
=\frac{\sum_{m=1}^M\gamma_m
\mathbb E_m\big[\ell_{\rm val}\{X,Y;\widehat\theta_m(\lambda)\}\mid\mathcal D_{\rm tr}\big]}
{\sum_{m=1}^M\gamma_m}.
\]
Since the oracle weights are unknown, we replace them with the training-based adaptive weights $\widehat\gamma_m(\lambda)$. Let $\widehat L_m^{\rm val}(\lambda)=n_{\rm val}^{-1}\sum_{i\in\mathcal S_m^{\rm val}}\ell_{\rm val}\{X_i,Y_i;\widehat\theta_m(\lambda)\}$ and $\widehat{\bar\gamma}(\lambda)=M^{-1}\sum_m\widehat\gamma_m(\lambda)$. The resulting weighted validation criterion is
\begin{equation}
\label{eq:cv:unpen:v2}
\widehat{\mathcal L}_{0}^{\rm val}(\lambda)
=\frac{\sum_{m=1}^M\widehat\gamma_m(\lambda)\widehat L_m^{\rm val}(\lambda)}
{M\widehat{\bar\gamma}(\lambda)},
\end{equation}
and ordinary CV selects $\widehat\lambda_n^{\rm cv}\in\arg\min_{\lambda\in\Lambda_n}\widehat{\mathcal L}_{0}^{\rm val}(\lambda)$.

However, the sampling variance of $\widehat{\mathcal L}_{0}^{\rm val}(\lambda)$ depends strongly on the concentration of the weights $\{\widehat\gamma_m(\lambda)\}_m$. {\blue If one client has a uniquely smallest training gradient norm, its normalized weight approaches one as $\lambda\to\infty$. In this case, the effective validation sample size approaches $n_{\rm val}$.} Consequently, a random downward fluctuation on that client can make \eqref{eq:cv:unpen:v2} artificially small and mislead ordinary CV into choosing an overly large $\lambda$.

To address this issue, we follow the uncertainty-penalization principle of \citet{maurer2009empirical} and add a penalty that grows as the weights become more concentrated. This leads to the pessimistic CV criterion
\begin{equation}
\label{eq:cv:pen:v2}
\widehat{\mathcal L}_{\tau}^{\rm val}(\lambda)
=\widehat{\mathcal L}_{0}^{\rm val}(\lambda)
+\frac{\tau}{\sqrt{n_{\rm val}}}
\frac{\{\sum_m\widehat\gamma_m(\lambda)^2\}^{1/2}}
{M\widehat{\bar\gamma}(\lambda)},
\end{equation}
which selects $\widehat\lambda_n^{\rm pes}\in\arg\min_{\lambda\in\Lambda_n}\widehat{\mathcal L}_{\tau}^{\rm val}(\lambda)$. Here $\tau\ge0$ controls the penalty strength, and $\tau=0$ recovers ordinary CV. When abnormal clients receive negligible weights, the penalty grows from roughly $\tau/\sqrt{|\mg|n_{\rm val}}$ under equal normal-client weights to $\tau/\sqrt{n_{\rm val}}$ under single-client dominance, so concentrated weights are penalized more heavily. To evaluate \eqref{eq:cv:pen:v2} without a central server, clients approximate the required averages via consensus iterations; the full DCV procedure is summarized in Algorithm~\ref{alg:dcv-pessimistic-v2}.

\begin{algorithm}[!t]
\setstretch{1}
\caption{DCV for Choosing $\lambda_n$}
\label{alg:dcv-pessimistic-v2}
\begin{algorithmic}[1]
\REQUIRE Candidate set $\Lambda_n$; training and validation sets $\{\mathcal S_m^{\rm tr},\mathcal S_m^{\rm val}\}_{m=1}^M$; penalty $\tau$; network matrix $W$; consensus rounds $T_{\rm cv}$.
\ENSURE Decentralized selectors $\{\widehat\lambda_{n,m}^{\rm DCV}\}_{m=1}^M$.
\FOR{each $\lambda\in\Lambda_n$}
\STATE Obtain $\{\widehat\theta_m(\lambda),\widehat\gamma_m(\lambda)\}_{m=1}^M$ using Algorithm~\ref{alg:multi_stage:aDFL} on $\mathcal D_{\rm tr}$.
\STATE Each client computes $\widehat L_m^{\rm val}(\lambda)$ and initializes $\boldsymbol z_m^{(0)}(\lambda)=\widehat\gamma_m(\lambda)\{\widehat L_m^{\rm val}(\lambda),1,\widehat\gamma_m(\lambda)\}^{\top}$.
\FOR{$t=0,\ldots,T_{\rm cv}-1$}
\STATE In parallel, update $\boldsymbol z_m^{(t+1)}(\lambda)=\sum_{k=1}^M w_{mk}\boldsymbol z_k^{(t)}(\lambda)$ for $1\le m\le M$.
\ENDFOR
\STATE Using the three components $z_{m,j}^{(T_{\rm cv})}(\lambda)$, client $m$ computes
\[
\widehat{\mathcal L}_{\tau}^{\rm DCV,m}(\lambda)
=\frac{z_{m,1}^{(T_{\rm cv})}(\lambda)
+\tau\{z_{m,3}^{(T_{\rm cv})}(\lambda)/(Mn_{\rm val})\}^{1/2}}
{z_{m,2}^{(T_{\rm cv})}(\lambda)}.
\]
\ENDFOR
\STATE Each client returns $\widehat\lambda_{n,m}^{\rm DCV}\in\arg\min_{\lambda\in\Lambda_n}\widehat{\mathcal L}_{\tau}^{\rm DCV,m}(\lambda)$.
\end{algorithmic}
\end{algorithm}

We next study the theoretical property of $\widehat\lambda_n^{\rm pes}$ with respect to the normal-client risk $\mathcal L_{\mg}^{\rm val}$. Denote the penalty term in \eqref{eq:cv:pen:v2} by $r_n(\lambda)=\tau\{\sum_m\widehat\gamma_m(\lambda)^2\}^{1/2}/\{\sqrt{n_{\rm val}}\sum_m\widehat\gamma_m(\lambda)\}$ and let $\lambda_{n,\mg}^\star\in\arg\min_{\lambda\in\Lambda_n}\mathcal L_{\mg}^{\rm val}(\lambda)$. Write $N_{\rm tr}=Mn_{\rm tr}$ and let $\widehat\theta^{\rm tr}$ be the whole-sample estimator computed on $\mathcal D_{\rm tr}$. Suppose all candidates share the same standard DFL initializers obtained from $\mathcal D_{\rm tr}$, initialized as in Proposition~\ref{prop:bad_estimator} and run until the optimization error is $O(\alpha+\SEW)$. We then need the following validation condition.

\begin{assumption}[Validation split]
\label{cond:cv-validation}
Conditional on $\mathcal D_{\rm tr}$, all validation observations are mutually independent and $0\le\ell_{\rm val}\{X,Y;\widehat\theta_m(\lambda)\}\le1$ almost surely for every $m$ and $\lambda\in\Lambda_n$.
\end{assumption}

Condition~\ref{cond:cv-validation} separates model fitting from validation via sample splitting \citep{stone1974cross,vandervaart2006oracle}, and boundedness can be enforced by clipping and rescaling with thresholds determined from $\mathcal D_{\rm tr}$. We then have the following result.

\begin{corollary}[Normal-client validation risk]
\label{cor:dcv-normal-risk}
Assume Conditions~\ref{ass:ps}--\ref{ass:bias} and \ref{cond:cv-validation} hold. Suppose $M\to\infty$, $M=o(n_{\rm tr}^4)$, $\varrho=o(1)$, and $\alpha+\SEW\lesssim N_{\rm tr}^{-1/2}$. The candidate range satisfies $C_0\log N_{\rm tr}\le\min\Lambda_n\le\max\Lambda_n=o\{(n_{\rm tr}/\log n_{\rm tr})^{1/2}\}$ for some $C_0>2/\flat_{\min}$. Further assume $(\max\Lambda_n)\|\widehat\theta^{\rm tr}-\theta_0\|=o_p(1)$. 
Set $\tau=\{\log(2KN)/2\}^{1/2}$. With probability tending to one,
\[
\mathcal L_{\mg}^{\rm val}(\widehat\lambda_n^{\rm pes})
-\mathcal L_{\mg}^{\rm val}(\lambda_{n,\mg}^\star)
\le 2r_n(\lambda_{n,\mg}^\star)
+o_p(1).
\]
\end{corollary}

Corollary~\ref{cor:dcv-normal-risk} {establishes the risk consistency of pessimistic CV}: the excess normal-client risk of $\widehat\lambda_n^{\rm pes}$ vanishes whenever $\log(2KN)/(Mn_{\rm val})=o(1)$, so pessimistic CV asymptotically matches the oracle candidate $\lambda_{n,\mg}^\star$ without knowing which clients are abnormal. {Moreover}, the leading bound $2r_n(\lambda_{n,\mg}^\star)$ depends only on the uncertainty at the optimal candidate $\lambda_{n,\mg}^\star$, rather than that of the selected candidate. {Under the conditions of Corollary~\ref{cor:dcv-normal-risk}, however, $\inf_{\lambda\in\Lambda_n}\widehat{\bar\gamma}(\lambda)\to_p1$, so all candidates in $\Lambda_n$ assign nearly uniform weights across normal clients.}

To study weight concentration, let $M_{\rm eff}(\lambda)=(\sum_m\widehat\gamma_m(\lambda))^2/\sum_m\widehat\gamma_m(\lambda)^2$ be the effective number of validation clients. Fix a sequence of training samples and suppose Condition~\ref{cond:cv-validation} holds. Consider a candidate $\lambda_1$ for which $M_{\rm eff}(\lambda_1)$ is bounded, while $M_{\rm eff}(\lambda)\ge cM$ for every other candidate and some fixed $c>0$. Let $\Delta_n$ be the conditional mean of \eqref{eq:cv:unpen:v2} at $\lambda_1$ minus the smallest such mean among the other candidates. Suppose that the conditional variance of each validation loss at $\lambda_1$ is bounded below by a fixed positive constant. If $n_{\rm val}\to\infty$, $\sqrt{n_{\rm val}}\Delta_n\to\kappa\in(0,\infty)$, and $\log(KN)/M\to0$, ordinary CV selects $\lambda_1$ with conditional probability bounded away from zero. With $\tau=\{\log(2KN)/2\}^{1/2}$, pessimistic CV avoids $\lambda_1$ and attains an excess weighted risk bounded by a constant times $\{\log(KN)/(Mn_{\rm val})\}^{1/2}$ with conditional probability tending to one.}

\section{Experiments}
\label{sec:exp}
In this section, we examine the finite-sample performance of the proposed aDFL method. We compare our aDFL algorithm with the following alternatives. 
\begin{itemize}[leftmargin=*]
\item \textbf{DFL} \citep{wu2023network}: 
The standard DFL algorithm without considering  abnormal clients.
The resulting DFL estimator is used as the initial value for aDFL. 
    \item \textbf{BRIDGE} \citep{fang2022bridge}: 
A robust method to aggregate neighbors' parameter estimates and run local gradient descent.
Coordinate-wise median (BRIDGE-M) and coordinate-wise trimmed mean (BRIDGE-T) are used for robustness.
\item\textbf{SLBRN} \citep{zhang2024119808high}: 
A robust gradient tracking algorithm, 
which aggregates both the parameter estimates and gradients from neighbors. 
Coordinate-wise median (SLBRN-M) and coordinate-wise trimmed mean (SLBRN-T) are used. 
\item\textbf{ClippedGossip} \citep{he2022selfcentered}:  
This algorithm 
utilizes ClippedGossip as the robust aggregation rule. It also uses
local momentum terms.
\end{itemize}
{In the synthetic-data experiments,} we adopt a fixed learning rate of $\alpha = 0.01$ and set the maximum number of iterations to $T = 500$ {for all algorithms}. For the aDFL algorithm, we implement the two-stage aDFL algorithm as detailed in { Algorithm~\ref{alg:multi_stage:aDFL}}. In addition, we apply the DCV algorithm in Section \ref{ap:subsec:cn} to select $\lambda_n$ from a candidate set of 5 grid points in the range $[\log(n)/25, \log(n)/5]$. For the BRIDGE-T and SLBRN-T algorithms, we use $\varrho$ as the trimming proportion for the trimmed mean operation. For the ClippedGossip algorithm, we employ the adaptive clipping strategy proposed by \citet{he2022selfcentered} to determine the clipping radius, with the hyperparameter $\delta_{\max} = 2\varrho$. Additionally, the momentum parameter is set to $0.9$ to align with that in \citet{he2022selfcentered}. 
To investigate the effect of the number of neighboring nodes, we further consider two different network structures. 
\begin{itemize}[leftmargin=*]
    \item \textbf{Directed Circle Network:} Assume that the clients are arranged in a fixed sequence with an in-degree $d_{m_1} = D> 0$ for each $1 \le m_1 \le M$. The network adjacency matrix $A=(a_{m_1m_2})$ is then defined with $a_{m_1 m_2} = 1$ if $m_2 = \big\{(m_1 + d - 1) \mod M \big\} + 1$ for $1 \le d \le D$, and $a_{m_1 m_2} = 0$ otherwise. Here, $a \mod b$ denotes the remainder when the integer $a$ is divided by the integer $b$. The resulting network structure should be of a circle type \cite{wu2023network}. 

    \item \textbf{Undirected Erd\H{o}s--R\'enyi Graph:} Consider an undirected Erd\H{o}s--R\'enyi graph represented by a symmetric adjacency matrix $ A = (a_{m_1m_2}) $, where $ a_{m_1m_2} = a_{m_2m_1} $ for all $ 1 \le m_1, m_2 \le M $. We generate each entry $ a_{m_1m_2} $ for $ 1 \le m_1 < m_2 \le M $ independently, with $ P(a_{m_1m_2} = 1) = q $ and $ P(a_{m_1m_2} = 0) = 1 - q $, where $ q \in (0, 1] $ is the link probability. Subsequently, we ensure symmetry by setting $ a_{m_2m_1} = a_{m_1m_2} $ for $ m_1 > m_2 $, and set $ a_{m_1m_2} = 0 $ for $ m_1 = m_2 $.

\end{itemize}

\subsection{Simulation Experiments on Synthetic Data}
\label{subsection:simu}
Following \citet{qian2024bymi}, we consider the linear regression model $Y_i = X_i^\top \theta_0 + \varepsilon_i$, where $\varepsilon_i\sim N(0,1)$ and $\theta_0 = (\mathbf{1}_s^\top, 0,\dots,0)^\top\in \mathbb{R}^p$ with $s = \lfloor 0.2 p \rfloor$.
For the distribution of $X_i$, we study two scenarios: a homogeneous scenario with $X_i \sim N_p(0,I_p)$, and a heterogeneous scenario in which each client generates $X_i$ from distinct multivariate normal distributions. 
Specifically, for each client $m$ ($1 \le m \le M$), feature vectors $X_i$s are generated from the multivariate normal distribution $N_p(\mu_m, \Sigma_m)$. Here, the mean vector $\mu_m \in \mathbb{R}^p$ is constructed by sampling each element independently from the uniform distribution $\mathcal{U}(-0.5, 0.5)$. The covariance matrix $\Sigma_m\in \mathbb{R}^{p \times p} $ is defined as $\Sigma_m = \big(\rho_m^{|j_1-j_2|}\big) $, where $\rho_m$ is sampled from $\mathcal{U}(0.2, 0.3)$. 
We consider the case where the data on abnormal clients is corrupted. 
{ Three types of data corruption are investigated:} (1) { \textbf{Bit-Flipping (BF)}}: The response variables $Y_i$'s on  abnormal clients are replaced by $\widetilde Y_i = - Y_i$. (2) { \textbf{Out-of-Distribution (OOD)}}: Features $X_i$'s on  abnormal clients are replaced by $\widetilde X_i = 0.7 X_i + V_p$, where entries of $V_p\in\mathbb{R}^p$ are independently generated from a uniform distribution $\mathcal{U}(0,1)$. 
(3) { \textbf{Model-Parameter Corruption (MP)}}: The parameters on abnormal clients are set as $\theta_c = (\mathbf{1}_{s_c}^\top,0,\dots,0)^\top \in\mathbb{R}^p$ with $s_c = \lfloor 0.1 p \rfloor$.   
We fix the feature dimension as $p=50$, the number of clients as $M=100$, and the local sample size as $n=100$. Thus, the total sample size is given by $N=M\times n = 10{,}000$. We randomly select $\lfloor \varrho M\rfloor$ clients as abnormal clients. 
We then use MSE on normal clients to assess the performance of estimators computed by different algorithms. Specifically, the MSE is defined as $\textup{MSE} = |\mathcal{A}^c|^{-1} \sum_{m\in\mathcal{A}^c} \|\wh\theta^{(m)}- \theta_0\|^2$, where $\wh\theta^{(m)}\in\mathbb{R}^p$ is the resulting estimator obtained on the $m$th client. 
For all algorithms, we replicate the experiments $20$ times in each setting. 

\begin{figure}[ht]
\centering
\includegraphics[width=0.75\textwidth]{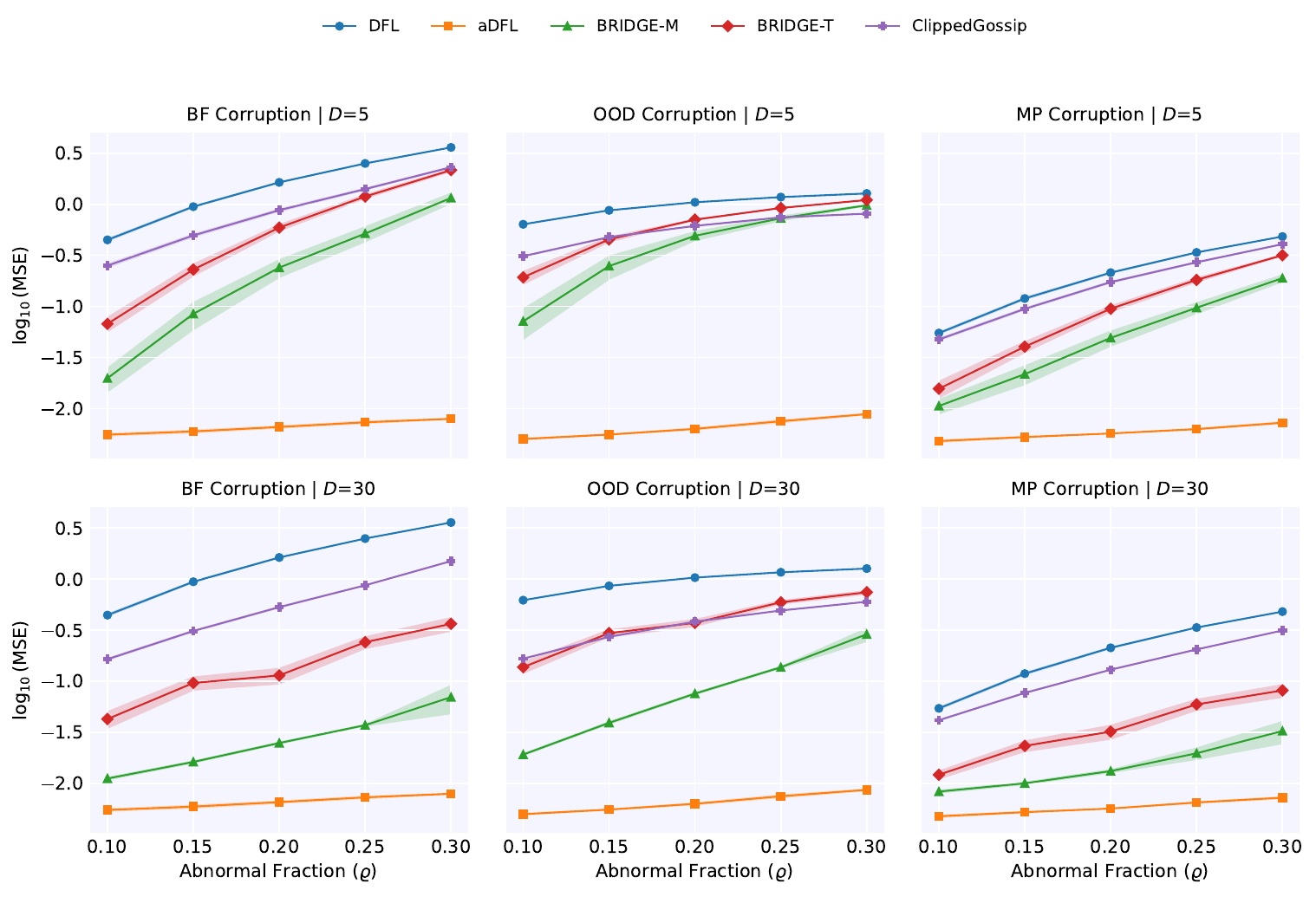}
\caption{The logarithm of MSE values versus the fraction of abnormal clients ($\varrho$) under the Directed Circle Network and the homogeneous scenario.   Different algorithms are evaluated under different corruption types and two in-degrees ($D$). {\blue  The shaded areas represent 95\% confidence bands based on 20 replications.}}
\label{fig:simu_net1}
\end{figure}

\begin{figure}[!t]
\centering\includegraphics[width=0.9\textwidth]{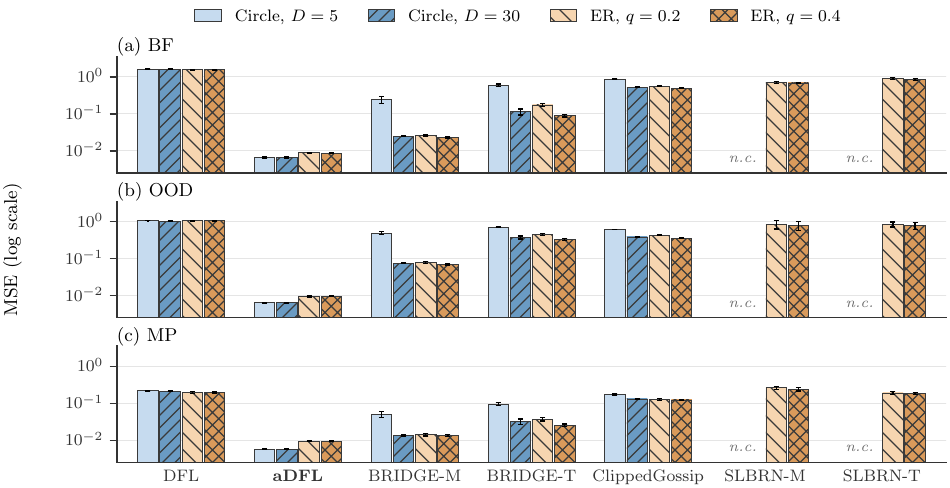}
\caption{{\blue MSE values (on the logarithmic scale) under different network structures in the homogeneous scenario with $\varrho=0.2$. Blue bars correspond to the Directed Circle Network with in-degree $D$, and orange bars correspond to the Undirected Erd\H{o}s--R\'enyi Graph with link probability $q$; lighter bars indicate the sparser network, and the four settings are further distinguished by fill patterns. Error bars are 95\% confidence intervals over 20 replications, and ``n.c.'' indicates that the algorithm fails to converge.}}
\label{fig:simu_network}
\end{figure}

\begin{figure}[!t]
\centering\includegraphics[width=0.9\textwidth]{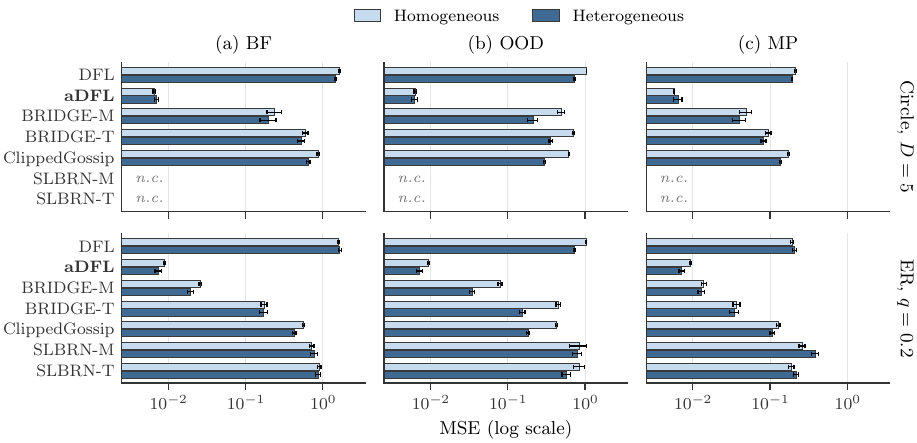}
\caption{{\blue MSE values (on the logarithmic scale) in the homogeneous and heterogeneous scenarios with $\varrho=0.2$, under the Directed Circle Network with $D=5$ (top) and the Undirected Erd\H{o}s--R\'enyi Graph with $q=0.2$ (bottom). Light and dark bars correspond to the homogeneous and heterogeneous scenarios, respectively; error bars and ``n.c.'' are as in Figure~\ref{fig:simu_network}.}}
\label{fig:simu_hetero}
\end{figure}

Figure~\ref{fig:simu_net1} plots the averaged MSE on the logarithmic scale, together with 95\% confidence bands, against $\varrho$ under the Directed Circle Network in the homogeneous scenario. To compare network structures, we summarize the results under both networks at $\varrho=0.2$ in Figure~\ref{fig:simu_network}. To assess the effect of data heterogeneity, we compare the homogeneous and heterogeneous scenarios at $\varrho=0.2$ in Figure~\ref{fig:simu_hetero}. 
Moreover, to further strengthen our simulation study, we explore additional experiments involving (1) two more realistic network structures, (2) a more complex data corruption type, and (3) a dynamic corruption scenario under specific settings. Across these settings, the results consistently demonstrate the robustness and effectiveness of our approach (results not shown).

Generally, the results under the two network structures show similar patterns, from which we obtain the following observations. First, under {the Directed Circle Network}, we find that the two SLBRN algorithms fail to converge, so the corresponding results are not reported. Second, as the abnormal fraction ($\varrho$) increases, the MSE of all algorithms increases significantly except for the aDFL algorithm. Furthermore, {\blue most} robust algorithms exhibit a smaller MSE compared to the standard DFL algorithm. Moreover, the aDFL algorithm achieves the smallest MSE among all these algorithms. 
Lastly, we find that various robust algorithms achieve lower MSE under the same data-corruption setting when $D$ increases from $5$ to $30$. 
This is expected because more information can be transmitted with a larger number of neighboring clients. {\blue  In addition, Figure~\ref{fig:simu_hetero} shows that aDFL remains the best-performing method in the heterogeneous scenario. Across all settings, its logarithmic MSE changes by at most 0.13 between the two scenarios, whereas the changes for the BRIDGE and ClippedGossip algorithms can exceed 0.35.}

\subsection{Application to Real Data}

In this section, we empirically evaluate the effectiveness of our proposed aDFL method on two classical datasets. 
(1) MNIST \citep{lecun1998gradient} consists of 70,000 handwritten digit images (0--9), with approximately 7,000 images per class. Among these, 60,000 are used for training and the remaining 10,000 are reserved for testing. For this dataset, we use the LeNet5 model with  $p = {\blue44,426}$  parameters. The initial values of the LeNet5 model are set using the Xavier uniform initializer. (2) CIFAR10 \citep{krizhevsky2009learning} comprises 60,000 color images, evenly distributed across 10 classes. Of these, 50,000 images are used for training and 10,000 are used for {testing}. For this dataset, we fine-tune the VGG16 model \citep{simonyan2014very} pre-trained on the ImageNet dataset,  with  $p = 5,130$  trainable parameters. 
In this experiment, we distribute all training data equally to $M=50$ clients. 
We consider two data distribution scenarios. The first is  \textbf{Homogeneous Scenario:} images from the entire training dataset are randomly and uniformly distributed across 50 clients.
The second is  \textbf{Heterogeneous Scenario:} images are first grouped according to their labels, with each label category evenly divided into $25$ subsets. These subsets are then assigned such that each client receives data from 5 subsets with different labels, ensuring that every client ultimately holds data associated with $5$ distinct labels.
For abnormal clients, we implement both OOD and label-flipping (LF) corruption. To be more precise, for \textbf{OOD}: the feature vectors $X_i$'s on abnormal machines are replaced by
$\widetilde{X_i} = 0.3 X_i + 3 {\blue Z_i}$, where {\blue$Z_i$ has the same shape as $X_i$ and its entries} are
independently generated from a standard normal distribution
$N(0, 1)$. For  { \textbf{Label-Flipping (LF)}}, we encode the image labels as numerical values ranging from 0 to 9. Subsequently, the response variables $Y_i$'s on the  abnormal machines are replaced by $\widetilde{Y_i} = (Y_i+1) \text{ mod } 10$. 

To speed up convergence, we adopt a constant-and-cut learning-rate scheduling strategy.
Specifically, for the MNIST dataset, a total of {\blue9,000} iterations are executed with an initial learning rate of $\alpha = 0.1$. The learning rate decreases to 0.05 after 200 iterations and to 0.01 after 4,000 iterations. For the CIFAR10 dataset, we run a total of 9,000 iterations with an initial learning rate of $\alpha = 1$. The learning rate is reduced to 0.5, 0.2, and 0.01 after 200, 5,000, and 8,000 iterations, respectively. 
For the proposed aDFL method, we recalculate the weights $\{\wh \gamma_m\}_m$ according to equation \eqref{eq:adaptive_weight} whenever the learning rate is adjusted.
In addition, we found that the ClippedGossip and SLBRN methods are highly sensitive to the learning rate in our setting. Therefore, we carefully adjusted the learning rate strategy for these methods separately. 

\begin{figure}[!ht]
  \centering
   \subfigure[MNIST Dataset]{%
   \includegraphics[width=0.75\textwidth]{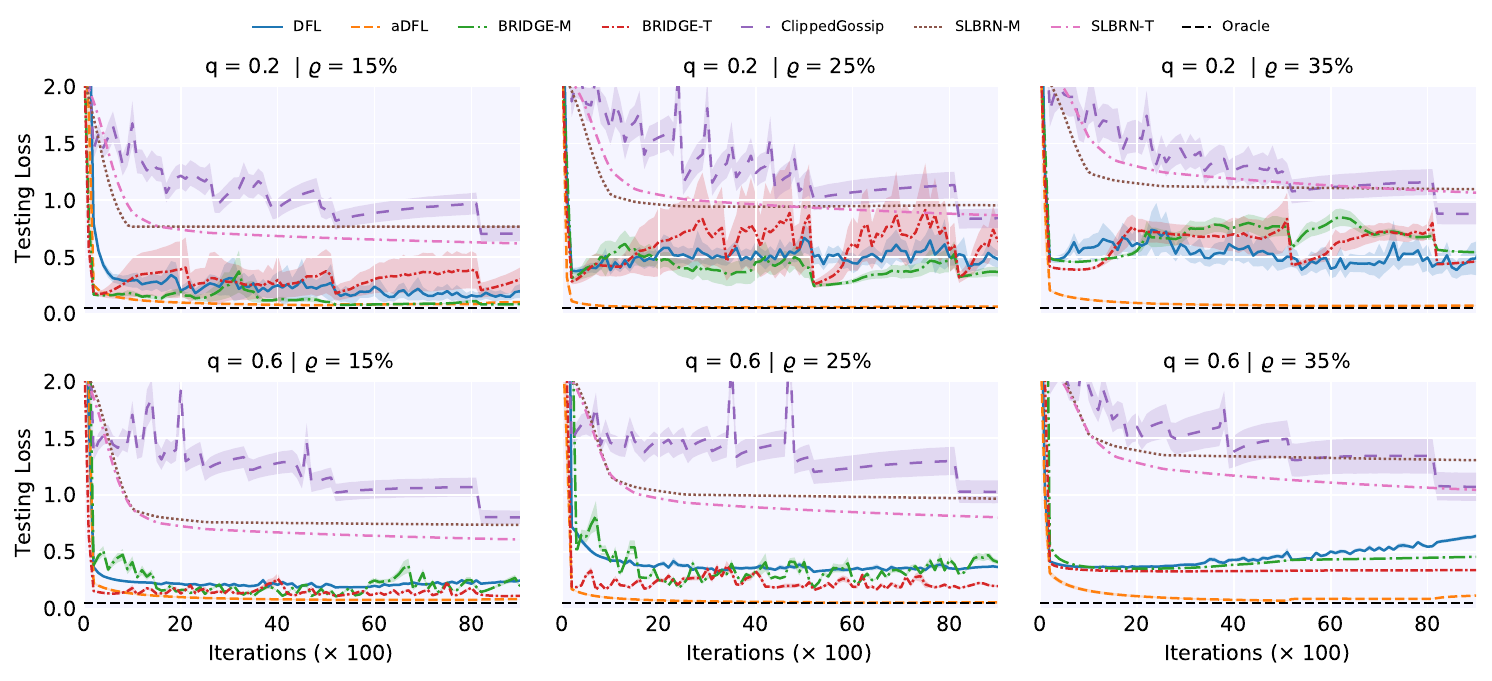}
     }
  \hfill
    \subfigure[CIFAR10 Dataset]{%
    \includegraphics[width=0.75\textwidth]{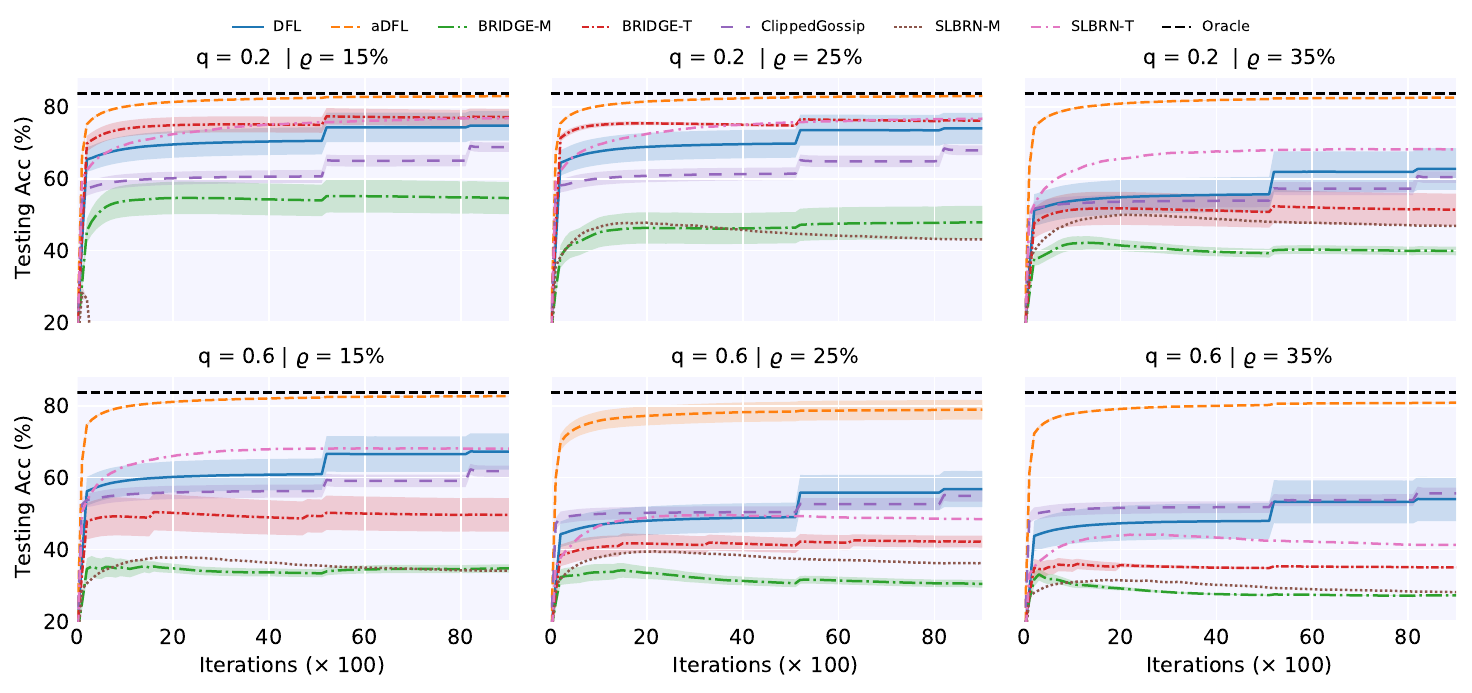}
  }
\caption{The testing loss/accuracy over iterations for two datasets in the heterogeneous scenario. Different methods are evaluated with varying link probabilities ($q$) and the fraction of abnormal clients ($\varrho$) under the LF corruption and Erd\H{o}s--R\'enyi Graph.}
\label{fig:CIFAR10}
\end{figure}

At the $t$th iteration, we evaluate the performance of the $m$th client on the testing set using loss and accuracy. 
We plot the averaged values and confidence bands of these results on normal clients. In addition to the competing methods discussed above, we include the oracle estimator as a reference.  { In the main text, we present results for MNIST and CIFAR10 under the heterogeneous scenario with LF corruption using the Erd\H{o}s--R\'enyi Graph;} see Figure \ref{fig:CIFAR10}.

From Figure \ref{fig:CIFAR10}, we find that as the fraction of abnormal clients increases or the number of neighbors decreases, the performances of competing methods decline significantly. 
Compared to competitors, our aDFL method achieves the best performance, which is comparable to that of the oracle across {\blue almost all} situations. 
This highlights aDFL's strong ability to be adaptive to different scenarios.

\FloatBarrier
\section{Conclusion}
\label{sec:con}
In this work, we propose aDFL, a robust decentralized federated learning method that dynamically adjusts each client's learning rate based on training behavior. { It preserves the original network topology and requires neither prior knowledge of client reliability nor a prescribed local contamination bound in the data-contamination setting.} We provide theoretical guarantees, and extensive experiments corroborate its effectiveness.
However, several limitations remain. First, 
{ the current design targets noisy or poisoned data; extending it to arbitrary message manipulation requires further study. Second, aDFL communicates every training round, which can be costly in large networks. Combining adaptive weighting with local updates is an important direction. {Third, although the DCV penalty controls sampling variability, concentrated weights can still shift the weighted validation risk away from the target risk when normal clients are heterogeneous. To mitigate this shift, future work may develop more robust criteria, such as evaluating all candidates with common weights or centering client-level losses.}}

\bibliographystyle{asa}
\bibliography{ref}
\end{document}